\documentclass[10pt,twocolumn,letterpaper]{article}

\usepackage{cvpr}              
\usepackage{booktabs}
\usepackage{multirow}
\usepackage[table]{xcolor}
\usepackage{colortbl}
\usepackage{subcaption}
\usepackage{tabularx}
\usepackage{algorithm}
\usepackage{algpseudocode}
\usepackage{bm}
\usepackage{graphicx}
\algnewcommand{\REQUIRE}{\item[\textbf{Require:}]}
\algnewcommand{\ENSURE}{\item[\textbf{Ensure:}]}
\algnewcommand{\TO}{\textbf{to}}
\algnewcommand{\OR}{\textbf{or}}

\definecolor{cvprblue}{rgb}{0.21,0.49,0.74}
\usepackage[pagebackref,breaklinks,colorlinks,allcolors=cvprblue]{hyperref}

\def\confName{CVPR}
\def\confYear{2026}

\title{Robust Remote Sensing Image–Text Retrieval with Noisy Correspondence }

\author{
Qiya Song\textsuperscript{1}, Yiqiang Xie\textsuperscript{1}, Yuan Sun\textsuperscript{2}\thanks{Corresponding authors.}, Renwei Dian\textsuperscript{3}, Xudong Kang\textsuperscript{3}
\\
\textsuperscript{1}Hunan Normal University, Changsha, China 410081\\
\textsuperscript{2}Sichuan University, Chengdu, China 610044\\  
\textsuperscript{3}Hunan University, Changsha, China 410082 \\
{\tt\small sqyunb@hnu.edu.cn, yiqiang\_xie@hunnu.edu.cn, sunyuan\_work@163.com}
}

\begin{document}
\maketitle
\begin{abstract}
As a pivotal task that bridges remote visual and linguistic understanding, Remote Sensing Image-Text Retrieval (RSITR) has attracted considerable research interest in recent years. However, almost all RSITR methods implicitly assume that image-text pairs are matched perfectly. In practice, acquiring a large set of well-aligned data pairs is often prohibitively expensive or even infeasible. In addition, we also notice that the remote sensing datasets (e.g., RSITMD) truly contain some inaccurate or mismatched image text descriptions. Based on the above observations, we reveal an important but untouched problem in RSITR, i.e., Noisy Correspondence (NC). To overcome these challenges, we propose a novel Robust Remote Sensing Image–Text Retrieval (RRSITR) paradigm that designs a self-paced learning strategy to mimic human cognitive learning patterns, thereby learning from easy to hard from multi-modal data with NC. Specifically, we first divide all training sample pairs into three categories based on the loss magnitude of each pair, i.e., clean sample pairs, ambiguous sample pairs, and noisy sample pairs. Then, we respectively estimate the reliability of each training pair by assigning a weight to each pair based on the values of the loss. Further, we respectively design a new multi-modal self-paced function to dynamically regulate the training sequence and weights of the samples, thus establishing a progressive learning process. Finally, for noisy sample pairs, we present a robust triplet loss to dynamically adjust the soft margin based on semantic similarity, thereby enhancing the robustness against noise. Extensive experiments on three popular benchmark datasets demonstrate that the proposed RRSITR significantly outperforms the state-of-the-art methods, especially in high noise rates. The code is available at: \url{https://github.com/MSFLabX/RRSITR}.

\end{abstract}    
\begin{figure}[!ht]
    \centering
    \begin{minipage}{0.49\linewidth}
        \centering
        \includegraphics[width=\textwidth,height=2.5cm]{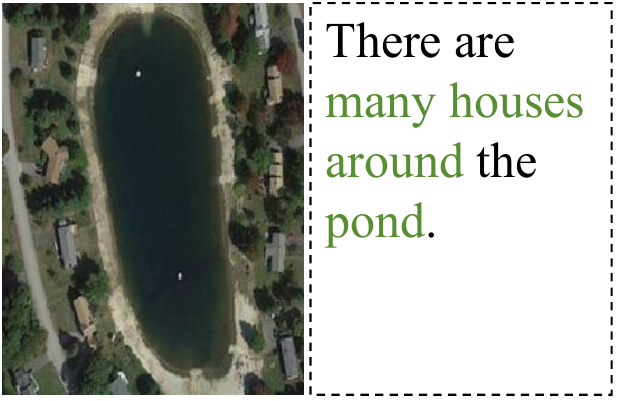}
        \caption*{\small (a) Perfect correspondence}
        \vspace{-5pt}
    \end{minipage}
    \hfill
    \begin{minipage}{0.49\linewidth}
        \centering
        \includegraphics[width=\textwidth,, height=2.5cm]{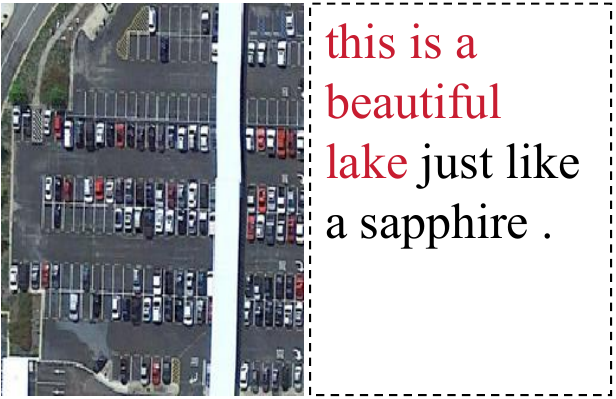}
        \caption*{\small (b) Noisy correspondence} \vspace{-5pt}
    \end{minipage}
    \caption{\small Representative examples of noise correspondence problems that really exist in the RSITMD dataset, wherein the sample pair is erroneously assumed to be semantically aligned. Since the model lacks explicit knowledge of which sample pairs are noisy during training, such incorrect supervision signals inevitably weaken retrieval performance. (a) and (b) represent clean sample pairs and noisy mismatched ones, respectively.}
    \label{fig1} 
    \vspace{-10pt}
\end{figure}
\section{Introduction}
Remote Sensing Image-Text Retrieval (RSITR) ~\cite{li2021image,zhou2025toward,wu2024spatial,yuan2022remote,tang2024prior} aims to bridge the semantic gap between aerial or satellite imagery and natural language descriptions, thereby enabling bidirectional querying, for example, retrieving relevant images given a textual query or retrieving descriptive captions for a given overhead scene. Thus, RSITR has attracted considerable research attention \cite{yuan2022lightweight,yuan2023parameter,zheng2023scale}, which demonstrates significant application potential in various domains such as environmental monitoring~\cite{li2020review}, urban planning~\cite{xiao2009review}, and disaster assessment~\cite{joyce2009review}. In recent years, numerous RSITR methods have been proposed, which could be primarily divided into two categories, i.e., Deep Convolutional Neural Network (DCNN) based methods and Transformer-based methods. The former employs convolutional neural networks and recurrent neural networks to extract and fuse image-text features for cross-modal retrieval. For instance, SIRS~\cite{zhu2024sirs} achieves fine-grained alignment via semantic segmentation, while MSA~\cite{yang2024transcending} employs multi-scale alignment for richer representations. The latter utilizes separate Transformer branches for each modality and aligns features in a shared space. Representative works include RemoteCLIP~\cite{liu2024remoteclip}, a foundation model trained with large-scale data excelling in retrieval tasks.

Although these methods have achieved pleasing performance, they rely on the ideal assumption of perfectly aligned image-text pairs in training data. However,  due to the predominantly nadir or vertical viewing geometry, remote sensing images lack human-centric egocentric visual priors (e.g., frontal or lateral views), leading to inherent ambiguity in text descriptions. Thus, constructing accurately the large scale of aligned remote sensing data is often prohibitively expensive and infeasible. As shown in Fig.\ref{fig1}, the RSITMD dataset factually contains some inaccurate/mismatched textual descriptions. Although a few studies~\cite{ji2024eliminate,pan2023prior,mikriukov2022unsupervised,tang2023interacting} have noticed the existence of meaningless or weakly correlated image-text pairs, almost all methods have failed to explicitly reveal such noisy pairs. Based on the above observation, we reveal an important but untouched problem in RSITR, i.e., Noisy Correspondence (NC). These mismatched pairs would inevitably mislead the learning process of the model, weakening the cross-modal retrieval performance. The core challenge is how to endow neural networks with robustness against such noise, thereby alleviating the interference of NC.

\begin{figure*}[!ht]
    \centering
    \includegraphics[scale=0.9, trim={0cm 0cm 0cm 0cm}]{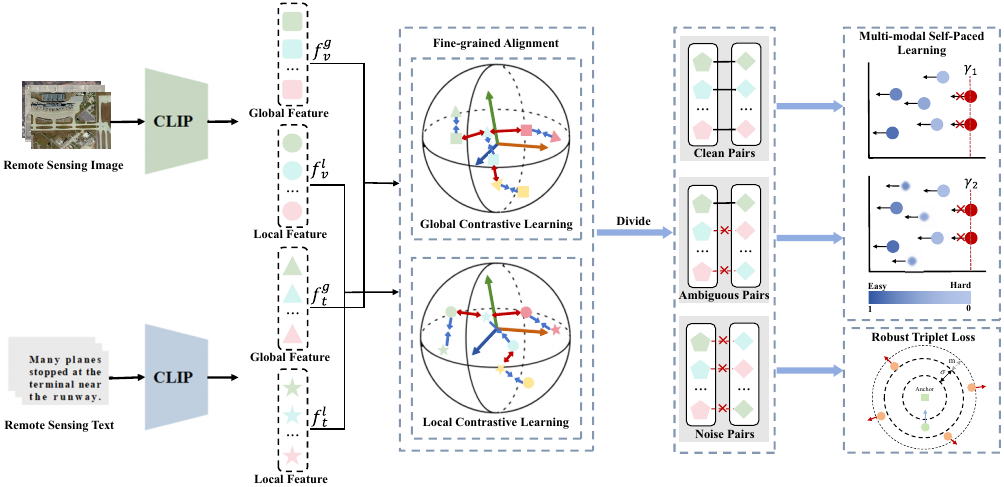}
    \caption{\small The overview of RRSITR. Specifically,  we first achieve fine-grained alignment by both global and local contrastive learning. Based on contrastive loss magnitude, training pairs are categorized into clean, ambiguous, and noisy. A multimodal self-paced learning strategy then dynamically weights clean and ambiguous pairs, enabling progressive learning from easy to hard correspondences. For noisy pairs, a robust triplet loss enhances model robustness and retrieval performance.
    }
    \label{fig2}\vspace{-15pt}
\end{figure*}

To alleviate the negative effects caused by NC, we propose a novel Robust Remote Sensing Image-Text Retrieval (RRSITR) paradigm that simulates the human cognitive ability to enable progressive learning from easy to hard instances for noisy multi-modal data, as shown in Fig.\ref{fig2}. Specifically, we first finely divide remote sensing data pairs into three groups according to the magnitude of loss, thereby enabling differentiated processing of the correspondence between clean, ambiguous, and noisy data. Then, we assign each pair a reliability weight derived from its loss value to achieve an adaptive assessment of its contribution to the learning process. Afterwards, for clean and ambiguous data, we design a new self-paced function to dynamically adjust both the weights and the learning sequence of the samples throughout training, thereby enabling the model to learn more semantic information from easy to hard. For noisy pairs, we propose a robust triplet loss to improve noise robustness by dynamically adjusting soft margins based on inter-sample semantic similarity. In summary, the main contributions of this work are as follows:
\begin{itemize}[itemsep=0pt]
\item We reveal and study a ubiquitous and untouched problem in
RSITR, dubbed noisy correspondence (NC). To the best of our knowledge, this paper could be the first work to prevent the model from being misled by incorrect visual semantic associations.
\item We propose a new multi-modal self-paced learning strategy to enable the model to learn from easy to hard in an organized manner for clean and ambiguous pairs. Moreover, we design a robust triplet loss to mitigate the adverse impact of NC.
\item Extensive experiments on three benchmark datasets show that our RRSITR significantly outperforms state-of-the-art methods, particularly in high noise scenarios.
\end{itemize}
\section{Related Work}
\label{s2}
\subsection{Remote Sensing Image-Text Retrieval}
Remote Sensing Image-Text Retrieval (RSITR) aims to retrieve semantically relevant remote sensing images given a text query, or vice versa. The core of this task lies in mapping data from different modalities into a shared embedding space, where cross-modal retrieval matching is achieved by maximizing the similarity of positive samples and minimizing that of negative samples. In recent years, a large number of RSITR methods have been proposed~\cite{zhang2023hypersphere,pan2023reducing,pan2023prior,yuan2022exploring,ji2023knowledge,zhu2024sirs}. For example, AMFMN~\cite{yuan2022exploring} proposes an asymmetric multimodal feature matching network to address the challenges posed by multi-scale features and redundant information in remote sensing images. However, AMFMN struggles to distinguish subtle differences in semantically similar descriptions. To address this, KAMCL~\cite{ji2023knowledge} introduces a knowledge-assisted learning framework that enhances the model's ability to discern fine-grained distinctions by leveraging discriminative information between similar descriptions. With the rapid development of vision-language pre-training models, some RSITR methods based on this paradigm have been proposed~\cite{gan2025semisupervised,wang2025cross,liu2024remoteclip,ji2024eliminate,chen2025context,zhang2024rs5m}. For instance, as the first vision-language foundation model in remote sensing, RemoteCLIP~\cite{liu2024remoteclip} could significantly enhance remote sensing retrieval performance, though it encounters challenges associated with high computational resource consumption. To alleviate this issue, CUP~\cite{wang2025cross} proposes a context and uncertainty-aware prompt network, significantly reducing resource overhead during model optimization. Although existing methods make considerable progress, they generally rely heavily on accurately paired remote sensing image-text data. However, in practical applications, constructing high-quality annotated datasets is not only costly and time-consuming but also inevitably leads to mismatched remote sensing image-text pairs. To address this, we reveal and study the NC problem in remote sensing image-text pairs and further propose a robust RSITR method to mitigate the negative impact of such NC on the retrieval performance.
\subsection{Self-paced Learning}
Inspired by the human cognitive learning process, self-paced learning (SPL)~\cite{kumar2010self} is proposed with a core idea that involves training models using a curriculum progressing from easy to hard samples. For instance, to mitigate the issue of noise or outliers, SCSM~\cite{liang2016self} employs an SPL strategy to learn gradually based on sample difficulty. However, this method requires manually designed weighting functions, which introduces additional hyperparameters and limits its applicability. To address this limitation, Meta-SPN~\cite{wei2021meta} achieves automatic learning of weight values. Nevertheless, most existing methods primarily leverage an easy to hard learning strategy to counteract feature noise, yet they typically consider difficulty only at the instance level. In response, DSCMH~\cite{sun2024dual} proposes an innovative dual-perspective self-paced learning mechanism that assesses difficulty from both instance-level and feature-level perspectives, thereby further enhancing model robustness. In contrast, DHaPH~\cite{huo2024deep} places greater emphasis on hard sample pairs, assigning them larger weights to excavate discriminative information. Furthermore, RSHNL~\cite{pu2025robust} focuses on handling multi-modal data with noisy labels to improve the noise robustness of hash models. Although these SPL approaches have achieved notable results, they predominantly address noise or outliers within the data. When dealing with remote sensing image-text data with NC, there remains a lack of research on leveraging SPL to mitigate the noisy correspondence problem.

\section{Method}\label{s3}
\subsection{Problem Formulation}\label{s31}
For clarity of presentation, in this paper, we define the image set $\mathcal{V}=\{I_{i}\}_{i=1}^{N_{v}}$ and the corresponding text set $\mathcal{T}=\{T_{i}\}_{i=1}^{N_{t}}$, where $N_{v}$ and $N_{t}$ representing the number of images and texts, respectively. The image-text pair is presented as $\mathcal{D}=\{(I_{i},T_{i},y_i)\}_{i=1}^{N}$, where each image $I_{i}$ is associated with a text $T_{i}$. $N$ represents the number of image-text pairs. $y_i \in \{0, 1\}$ represents the correspondence label, where $y_i = 1$ indicates the correct correspondence, otherwise $y_i = 0$. Following~\cite{ji2024eliminate}, for the input image-text data $\{(I_{i},T_{i}\}$, we employ the CLIP visual encoder to extract its global visual feature $f_{v}^{g}$ and local visual features $\{f_{v}^{1},\ldots, f_{v}^{d_{1}}\}$. For an input text $T_{i}$, we use the CLIP text encoder to extract its global text feature $f_{t}^{g}$ and local text features $\{f_{t}^{1},\ldots, f_{t}^{d_{2}}\}$. The above process could be formulated as:
\begin{equation}   
\footnotesize
\begin{aligned}
f_{v}^{g},f_{v}^{1}...f_{v}^{d_{1}}=\varphi(I_i),\\
f_{t}^{g},f_{t}^{1}...f_{t}^{d_{2}}=\phi(T_i), 
\end{aligned}
\end{equation}
where $\varphi$ and $\phi$ are the CLIP visual encoder and the CLIP text encoder, respectively. $d_{1}$ and $d_{2}$ are the number of local visual features and local text features, respectively.

\subsection{Fine-grained Alignment}
\label{s32}
To achieve more precise cross-modal alignment between images and texts, we propose a fine-grained alignment scheme to leverage both global information and local details. Specifically, we first calculate the global similarity of each image-text pair by the cosine similarity, i.e.,
\begin{equation} \footnotesize
\begin{aligned}
S^{g} = \cos\left( f_{v}^{g}, f_{t}^{g} \right).
\end{aligned}
\end{equation}
Afterwards, we compute the local similarity of each local block $S^l_{ij}=cos(f_v^i,f_t^j)$. Then, we compute the local similarity of each image-text pair by performing two consecutive L2-norm operations, i.e.,
\begin{equation} \footnotesize
\begin{aligned}
S^{l} = \parallel S^l_{ij} \parallel_{2,2},
\end{aligned}
\end{equation}
where $\parallel \cdot \parallel_{2,2}$ represents two consecutive L2-norm.  

To achieve the fine-grained alignment of multi-modal data, we adopt the InfoNCE loss~\cite{oord2018representation} to maximize the similarities of positive pairs while minimizing the similarities of the negative pairs, which could be defined as:
\begin{equation}   
\footnotesize 
 \mbox{\tiny $\displaystyle
	\begin{aligned}
	\mathcal{L}_{info}^{g} = -\frac{1}{N}\sum_{j=1}^{N}\left(\log\tfrac{\exp\left(S_{jj}^g/\tau\right)}{\sum_{i=1}^{N}\exp\left(S_{ji}^{g}/\tau\right)} + \log\tfrac{\exp\left(S_{jj}^g/\tau\right)}{\sum_{i=1}^{N}\exp\left(S_{ij}^{g}/\tau\right)}\right),\\
	\mathcal{L}_{info}^{l} = -\frac{1}{N}\sum_{j=1}^{N}\left(\log\tfrac{\exp\left(S_{jj}^{l}/\tau\right)}{\sum_{i=1}^{N}\exp\left(S_{ji}^{l}/\tau\right)} + \log\tfrac{\exp\left(S_{jj}^{l}/\tau\right)}{\sum_{i=1}^{N}\exp\left(S_{ij}^{l}/\tau\right)}\right),
	\end{aligned}
    $}
\end{equation}
\noindent where $S^g_{jj}$ and $S^l_{jj}$ denote the $j$-th global and local positive sample pairs, respectively. $\sum_{i = 1}^{N}S_{ji}^{g}$ and $\sum_{i = 1}^{N}S_{ji}^{l}$ represent the row-wise sums of the similarity matrices for image-to-text alignment. $\sum_{i = 1}^{N}S_{ij}^{g}$ and $\sum_{i = 1}^{N}S_{ij}^{l}$ represent the row-wise sums of the similarity matrices for text-to-image alignment. $\tau$ denotes the temperature hyperparameter, and $N$ denotes the batch size. Finally, we apply the InfoNCE loss to achieve both global and local alignment, which could be expressed as:
\begin{equation}   \footnotesize
	\begin{aligned}
	\mathcal{L}_{info}^{gl} = \mathcal{L}_{info}^{g} + \mathcal{L}_{info}^{l},
	\end{aligned}
\end{equation}
where $\mathcal{L}_{info}^{g}$ is the global contrastive loss and $\mathcal{L}_{info}^{l}$ is the local contrastive loss.

\subsection{Multi-modal Self-Paced Learning} 
\label{s33}

Inspired by the human cognitive process of learning from easy to hard, we propose a multi-modal self-paced learning scheme to mitigate the interference of NC during the training. Specifically, RRSITR dynamically partitions all training sample pairs into three subsets, i.e., clean sample pairs, ambiguous sample pairs, and noisy sample pairs, based on their total contrastive loss values. For samples of different qualities, we design a self-paced weighting function to dynamically assess sample difficulty and control their participation order and contribution during training. The model is trained to first learn from simple and reliable clean samples along with ambiguous samples, then gradually incorporate more challenging ones. This progressive curriculum learning strategy enables the model to steadily capture the alignment between images and texts, thereby effectively suppressing the negative impact of incorrect correspondences and enhancing overall robustness. RRSITR dynamically adjusts the participation manner and weight of sample pairs based on the relationship between the current loss $\ell_{i}$ and thresholds $\gamma_1$, $\gamma_2$.

When $\ell_{i} < \gamma_1$, clean sample pairs are deemed reliable and easy to learn. RRSITR incorporates these samples in the early stages of training and gradually introduces more challenging ones. When $\gamma_1 \leq \ell_{i} < \gamma_2$, ambiguous pairs are considered relatively reliable but exhibit some degree of difficulty. RRSITR prioritizes the learning of these samples. The loss functions for these samples are defined as:
\begin{equation}   \footnotesize
\begin{aligned}
\mathcal{L}_{S1}=\frac{1}{N}\sum_{i=1}^{N}w_{i}(\underbrace{\mathcal{L}_{i}^{g}+\mathcal{L}_{i}^{l})}_{{\ell_{\mathrm{i}}}}+\frac{1}{N}\sum_{i=1}^{N}\mathcal{R}\left(w_{i},\gamma_1\right),\\
\mathcal{L}_{S2}=\frac{1}{N}\sum_{i=1}^{N}w_{i}(\underbrace{\mathcal{L}_{i}^{g}+\mathcal{L}_{i}^{l})}_{{\ell_{\mathrm{i}}}}+\frac{1}{N}\sum_{i=1}^{N}\mathcal{R}\left(w_{i},\gamma_2\right),
\label{L_s}
\end{aligned}
\end{equation}
\noindent where $\ell_{i}$ denotes the sum of the global contrastive loss and the local contrastive loss for the $i$-th sample pair, $w_{i} \in [0, 1]$ denotes the importance weight of the $i$-th sample pair, assessing its reliability, and $\mathcal{R}(w_i, \gamma)$ is a self-paced regularizer governed by the learning pace parameter $\gamma$, assigning a weight $w_i$ to each sample based on its learning difficulty.

When $\ell_{i} \geq \gamma_2$, the sample pairs are considered potentially noisy correspondence. Thus, the weight $w_{i}$ is set to zero to exclude it from training. In summary, RRSITR simulates the human cognitive learning process that progresses from easy to hard, gradually incorporating more sample pairs into the model training. 

To achieve finer control over the sample selection strategy and training progress for more robust learning with noisy correspondence, we design a novel self-paced regularizer $\mathcal{R}(w_i, \gamma)$, expressed as follows:
\begin{equation}   \footnotesize
\mbox{\scriptsize $\displaystyle
{R}(w_i,\gamma)=\left\{
        \begin{aligned}
            & -\frac{2}{\pi}\gamma\left[w_iarccos(w_i)-\sqrt{1-{w_i}^2}\right], && \ell_i < \gamma, \\
            &0, && \ell_i \geq \gamma.
        \end{aligned}
    \right.
    $}
    \label{R_w}
\end{equation}
In general, the weight could be viewed as an indicator of the easiness for each noisy pair. A higher weight suggests the instance is simpler. As the learning process proceeds, the loss gradually decreases, thereby increasing the weight. 

\subsection{Robust Triplet Loss}
The hard triplet loss is a widely adopted metric learning objective in cross-modal representation learning~\cite{faghri2017vse++,zhang2022negative,pan2023fine,lee2018stacked,diao2021similarity}. It enforces semantic alignment by encouraging the similarity between a positive cross-modal pair to exceed that of negative pairs by a margin. Hard triplet loss achieves strong empirical performance by leveraging hard negative mining, particularly the hardest negatives within each mini-batch. However, in the presence of noisy data, positive sample pairs may be incorrectly matched, resulting in a low similarity score $S^{g}(\mathbf{v}_{i},\mathbf{t}_{i})$. Under such circumstances, the difference between the positive sample similarity and the hardest negative sample similarity often increases, thereby expanding the margin and forcing the model to distinguish between positive and negative samples with stricter standards, effectively suppressing noise interference. In contrast, fixed-margin methods often lead the model to become overconfident on easy samples while underlearning hard samples. For noisy pairs, we propose a Robust Triplet Loss (RTL) to further enhance robustness against NC through an adaptive margin mechanism and a hard sample mining strategy. The loss function could be written as follows:
\begin{equation}   \footnotesize
\begin{aligned}
\mathcal{L}_{soft}& =\frac{1}{N}\sum_{i=1}^{N}([\hat{\mu}_{i}-S^{g}(\mathbf{v}_{i},\mathbf{t}_{i})+S^{g}(\mathbf{v}_{i},\hat{\mathbf{t}}_{h})]_{+} \\
 & +[\hat{\zeta}_{i}-S^{g}(\mathbf{v}_{i},\mathbf{t}_{i})+S^{g}(\hat{\mathbf{v}}_{h},\mathbf{t}_{i})]_{+}),
\end{aligned}
\end{equation}
where $S^{g}(\mathbf{v}_{i},\mathbf{t}_{i})$ denotes the global similarity of positive sample pairs and $[x]_+\equiv\max(x,0)$. Here, $\hat{v}_h = \arg max_{v_j \neq v_i}S^{g}(v_j, t_i)$ and $\hat{t}_h = \arg max_{t_j \neq t_i}S^{g}(v_i, t_j)$ are the most similar negatives to $t_i$ and $v_i$ in a batch. The soft margins $\hat{\mu}_{i}$ and $\hat{\zeta}_{i}$ are adaptively determined as follows:
\begin{equation}   \footnotesize
\begin{aligned}& \hat{\mu}_{i}=\sigma*(1+\max(0,S^{g}(\mathbf{v}_{i},\hat{\mathbf{t}}_{h})-S^{g}(\mathbf{v}_{i},\mathbf{t}_{i}))), \\& \hat{\zeta}_{i}=\sigma*(1+\max(0,S^{g}(\hat{\mathbf{v}}_{h},\mathbf{t}_{i})-S^{g}(\mathbf{v}_{i},\mathbf{t}_{i}))),
\end{aligned}
\end{equation}
where $\sigma>0$ is a base margin hyperparameter. Through dynamic margin adjustment, RTL balances the learning focus, preventing the model from overfitting on noisy data. Furthermore, the bidirectional symmetric treatment in the loss function ensures that noise in both modalities can be covered by the adaptive margin mechanism, enhancing overall robustness and image-text consistency.

\subsection{Overall Objective and Inference}
The overall objective of our RRSITR is defined as follows:
\begin{equation}   \footnotesize 
	\begin{aligned}
\mathcal{L}_{overall}=\mathcal{L}_{S1}+\lambda_1\mathcal{L}_{S2}+\lambda_2\mathcal{L}_{soft},
	\end{aligned}
\end{equation} 
where $\lambda_1$ and $\lambda_2$ are the balancing factors. Due to the limited space, we attach the detailed training process is provided in the supplementary materials.

In the inference stage, to obtain a more fine-grained similarity relationship, we adopt a weighted scheme to fuse the global and local similarities, thereby obtaining the fine-grained image-text similarity $S^{f}$ as follows:  
\begin{equation}   \footnotesize
	\begin{aligned}
	S^{f} = \alpha S^{g} +(1-\alpha)S^{l},
	\end{aligned}
\end{equation}
where $\alpha$ is a hyperparameter.

\subsection{Weight Analysis}
\label{s34}
To solve the minimization problem in \Cref{L_s}, we conduct a thorough analysis of how the weights $w_i$ influence the loss function $\mathcal{L}_S$. Specifically, we alternately update the weight $w_i$ and the network parameter $\Theta$ to minimize $\mathcal{L}_S$. Under the condition of the fixed $\Theta$, the optimal solution of the weight $w_i$ could be written as:
\begin{equation}
\footnotesize 
\scalebox{0.85}{$
\displaystyle
{w}_{i} =\left\{
        \begin{aligned}
            & \mathop{\mathrm{argmin}}_{w_i \in [0,1]} \left\{ w_i \ell_i - \frac{2}{\pi}\gamma \left[ w_i \arccos(w_i) - \sqrt{1 - w_i^2} \right] \right\}, && \ell_i < \gamma, \\
            & 0, && \ell_i \geq \gamma.
        \end{aligned}
    \right. 
$}
\label{w_1}
\end{equation}
When $\ell_i - \gamma < 0$, let the derivative of \Cref{w_1} to zero, we can obtain:
\begin{equation}   \footnotesize  
w_i^* = \cos\left( \frac{\pi}{2} \cdot \frac{\ell_i}{\gamma} \right).
\label{eq14}
\end{equation}
Given that $\gamma \geq 0$ and $\ell_i \geq 0$, it can be derived that $w_i \in [0, 1]$. Thus, the optimal weight solution is expressed as follows:
\begin{equation}   \footnotesize 
	w_i^* =
    \left\{
        \begin{aligned}
            &\cos\left(\frac{\pi}{2} \cdot \frac{\ell_i}{\gamma}\right), && \ell_i < \gamma, \\
            &0, && \ell_i \geq \gamma.
        \end{aligned}
    \right.
    \label{w_2}
\end{equation}
When the loss value $\ell_i$ is excessively large (i.e., $\ell_i \geq \gamma$), the corresponding sample pairs are treated as hard data with noisy correspondences and are assigned a weight of zero. When $\ell_i < \gamma$, the $i$-th sample pair with larger weight is implicitly regarded as easy, whereas those with smaller weights are regarded as hard. Overall, this strategy not only effectively distinguishes noise pairs but also enhances the model's robustness and generalization capability.

\begin{table*}[htb]
    \scriptsize 
    \setlength{\extrarowheight}{-3pt} 
    \centering
    \caption{Image-text retrieval performance under different noise ratios on the RSITMD dataset.}
\begin{tabular*}{\textwidth}{@{\extracolsep{\fill}}cccccccccc@{}}
\toprule
\multirow{2}{*}{Noise} & \multirow{2}{*}{Method} &\multirow{2}{*}{Ref.} & \multicolumn{3}{c}{Image-to-Text Retrieval} & \multicolumn{3}{c}{Text-to-Image Retrieval} & \multirow{2}{*}{mR}  \\
\cmidrule(lr){4-6} \cmidrule(lr){7-9}
 & & & R@1 & R@5 & R@10 & R@1 & R@5 & R@10
 \\
\midrule
\multirow{11}{*}{20\%} 
 & AMFMN&TGRS'22 & 9.56±1.17 & 26.90±2.02 & 39.29±1.96 & 7.66±0.78 & 31.19±0.49 & 51.42±0.94 & 27.67±0.93 \\
  & HVSA&TGRS'23 & 7.87±0.99 & 23.54±1.52 & 34.87±2.72 & 6.35±0.29 & 27.57±0.71 & 47.46±1.03 & 24.61±0.74 \\
  & KAMCL&TGRS'23 & 10.80±0.89 & 27.17±1.08 & 39.87±1.32 & 9.59±0.61 & 34.04±0.65 & 51.23±1.15 & 28.78±0.76 \\
  & SWAN&ICMR'23 & 9.29±2.28 & 24.25±1.38 & 37.43±2.55 & 7.44±0.45 & 30.93±0.84 & 52.71±0.92 & 27.01±0.81 \\
  & PIR&ACMMM'23 & 10.09±1.66 & 29.25±1.69 & 44.95±1.19 & 8.74±0.89 & 33.45±1.25 & 54.04±1.06 & 30.09±0.89 \\
  & S-CLIP&NeurIPS'23 & 6.99±0.86 & 25.66±0.31 & 41.06±1.98 & 6.90±1.01 & 24.87±2.68 & 38.85±1.60 & 24.05±1.20 \\
  & SIRS&TGRS'24 & 10.13±0.68 & 28.27±0.87 & 40.84±0.89 & 7.70±0.62 & 32.30±0.78 & 52.87±0.78 & 28.69±0.14 \\
  & MSA&TGRS'24 & 12.88±0.88 & 31.33±1.46 & 45.49±2.33 & 10.11±0.28 & 38.14±1.48 & 57.64±0.83 & 32.60±0.54 \\
  & SEMICLIP&ICLR'25 & 8.40±1.15 & 27.98±0.83 & 42.48±1.70 & 7.34±1.50 & 28.58±2.03 & 45.10±1.71 & 26.65±1.13 \\
  & CUP&TNNLS'25 & \underline{19.78±0.64} & \underline{38.94±2.30} & \underline{52.52±1.28} & \underline{15.30±1.10} & \underline{44.64±1.53} & \underline{62.73±0.97} & \underline{38.99±1.16} \\
  & RRSITR&Ours & \textbf{24.60±1.38} & \textbf{46.68±1.11} & \textbf{58.85±1.92} & \textbf{20.19±0.89} & \textbf{53.04±0.70} & \textbf{70.12±0.37} & \textbf{45.58±0.38} \\
\midrule
\multirow{11}{*}{40\%} 
 & AMFMN&TGRS'22 & 7.30±0.73 & 21.72±0.81 & 33.89±2.03 & 6.13±0.67 & 27.04±0.83 & 45.60±0.84 & 23.61±0.60 \\
  & HVSA&TGRS'23 & 6.33±0.72 & 18.36±1.20 & 28.80±2.48 & 4.85±0.31 & 19.87±0.98 & 36.48±1.81 & 19.12±1.08 \\
  & KAMCL&TGRS'23 & 5.62±2.37 & 16.24±6.38 & 26.10±9.99 & 4.99±1.68 & 20.03±5.83 & 33.12±7.94 & 17.68±5.55 \\
  & SWAN&ICMR'23 & 7.26±0.80 & 22.39±0.65 & 35.53±0.55 & 6.27±0.17 & 27.80±0.65 & 48.71±1.53 & 24.66±0.40 \\
  & PIR&ACMMM'23 & 7.66±0.83 & 24.87±1.47 & 37.43±0.95 & 6.74±0.53 & 29.10±0.91 & 50.82±0.42 & 26.10±0.50 \\
  & S-CLIP&NeurIPS'23 & 3.36±1.45 & 12.70±4.86 & 20.97±8.01 & 3.05±1.49 & 11.33±5.65 & 19.82±8.69 & 11.87±4.99 \\
  & SIRS&TGRS'24 & 7.66±0.89 & 22.61±0.53 & 35.35±1.05 & 5.97±0.64 & 26.86±0.79 & 44.85±0.85 & 23.88±0.51 \\
  & MSA&TGRS'24 & 3.54±0.90 & 12.08±2.98 & 21.11±4.79 & 5.24±0.62 & 22.35±2.23 & 37.53±3.41 & 16.98±2.06 \\
  & SEMICLIP&ICLR'25 & 6.19±1.15 & 21.59±1.24 & 34.11±1.12 & 6.09±0.83 & 22.10±1.87 & 34.52±2.00 & 20.77±1.24 \\
  & CUP&TNNLS'25 & \underline{17.97±0.60} & \underline{37.25±1.07} & \underline{51.19±0.87} & \underline{13.72±0.69} & \underline{40.91±0.72} & \underline{59.35±0.35} & \underline{36.73±0.26} \\
  & RRSITR&Ours & \textbf{23.23±1.53} & \textbf{43.63±0.43} & \textbf{56.15±0.38} & \textbf{19.27±0.88} & \textbf{51.09±1.10} & \textbf{68.48±0.64} & \textbf{43.64±0.28} \\
\midrule
\multirow{11}{*}{60\%} 
 & AMFMN&TGRS'22 & 6.46±1.04 & 20.27±1.56 & 31.28±1.80 & 5.20±1.03 & 21.76±1.20 & 37.33±1.72 & 20.38±1.02 \\
  & HVSA&TGRS'23 & 4.78±0.48 & 14.20±1.59 & 22.92±1.69 & 3.17±0.69 & 15.79±1.05 & 27.21±1.09 & 14.68±0.71 \\
  & KAMCL&TGRS'23 & 0.44±0.24 & 1.73±0.86 & 3.59±1.42 & 0.59±0.28 & 3.15±1.45 & 5.96±2.57 & 2.58±0.80 \\
  & SWAN&ICMR'23 & 6.42±0.89 & 19.43±1.17 & 32.17±0.92 & 5.51±0.08 & 24.27±0.97 & 43.62±1.26 & 21.90±0.57 \\
  & PIR&ACMMM'23 & 7.39±1.12 & 23.10±0.89 & 36.68±1.08 & 5.38±0.62 & 25.46±0.85 & 46.23±1.15 & 24.04±0.66 \\
  & S-CLIP&NeurIPS'23 & 0.71±0.38 & 2.74±0.75 & 4.33±1.09 & 0.40±0.16 & 2.30±0.86 & 3.94±0.88 & 2.40±0.56 \\
  & SIRS&TGRS'24 & 6.24±0.36 & 19.56±0.82 & 29.82±0.96 & 4.24±0.29 & 18.37±0.45 & 31.41±0.49 & 18.27±0.15 \\
  & MSA&TGRS'24 & 0.26±0.09 & 1.64±0.55 & 3.32±0.81 & 1.68±0.32 & 7.58±0.44 & 14.50±1.46 & 4.83±0.50 \\
  & SEMICLIP&ICLR'25 & 3.89±0.69 & 13.41±1.73 & 21.25±2.80 & 3.23±0.63 & 12.87±1.98 & 21.17±2.23 & 12.64±1.45 \\
  & CUP&TNNLS'25 & \underline{15.62±0.65} & \underline{34.25±1.86} & \underline{48.14±1.32} & \underline{11.65±0.73} & \underline{37.37±2.00} & \underline{55.06±2.66} & \underline{33.68±0.83} \\
  & RRSITR&Ours & \textbf{22.03±0.59} & \textbf{41.50±1.60} & \textbf{54.20±1.06} & \textbf{17.72±0.66} & \textbf{49.63±0.81} & \textbf{67.30±0.63} & \textbf{42.06±0.28} \\
\midrule
\multirow{11}{*}{80\%} 
 & AMFMN&TGRS'22 & 3.50±0.26 & 13.54±0.68 & 20.89±1.23 & 3.69±0.35 & 13.81±0.59 & 24.61±0.31 & 13.34±0.47 \\
  & HVSA&TGRS'23 & 2.83±0.69 & 8.85±1.24 & 14.56±1.34 & 1.90±0.24 & 8.02±0.51 & 14.44±0.70 & 8.43±0.55 \\
  & KAMCL&TGRS'23 & 0.35±0.18 & 1.02±0.46 & 2.16±0.45 & 0.46±0.15 & 1.79±0.21 & 3.37±0.35 & 1.53±0.17 \\
  & SWAN&ICMR'23 & 3.23±1.39 & 12.08±1.89 & 20.80±0.90 & 3.33±0.48 & 14.34±1.37 & 26.16±1.78 & 13.32±1.10 \\
  & PIR&ACMMM'23 & 3.67±0.72 & 13.63±1.69 & 23.41±1.72 & 2.58±0.25 & 12.18±0.28 & 22.03±0.74 & 12.92±0.66 \\
  & S-CLIP&NeurIPS'23 & 0.26±0.16 & 1.50±0.47 & 2.70±0.33 & 0.26±0.16 & 1.24±0.41 & 2.65±0.75 & 1.44±0.30 \\
  & SIRS&TGRS'24 & 3.18±0.57 & 10.31±0.97 & 17.03±1.96 & 2.31±0.42 & 9.35±0.56 & 15.87±0.71 & 9.68±0.71 \\
  & MSA&TGRS'24 & 0.26±0.09 & 1.28±0.72 & 2.56±0.57 & 0.86±0.29 & 3.74±0.49 & 7.51±0.89 & 2.70±0.20 \\
  & SEMICLIP&ICLR'25 & 1.16±0.44 & 5.17±1.10 & 8.75±1.61 & 1.39±0.65 & 5.50±1.15 & 8.80±1.04 & 5.13±0.92 \\
  & CUP&TNNLS'25 & \underline{10.71±0.75} & \underline{27.61±1.43} & \underline{40.35±1.38} & \underline{8.27±0.78} & \underline{30.38±1.47} & \underline{48.22±2.67} & \underline{27.59±0.99} \\
  & RRSITR&Ours & \textbf{16.90±1.49} & \textbf{33.98±1.84} & \textbf{46.82±1.75} & \textbf{13.72±1.12} & \textbf{42.70±1.08} & \textbf{61.47±0.69} & \textbf{35.93±0.68} \\
\bottomrule
\end{tabular*}
    \label{tab_rsitmd}\vspace{-12pt}
\end{table*}

\section{Experiments}\label{s4}

\begin{table*}[!htb]
    \scriptsize 
    \setlength{\extrarowheight}{-1pt} 
    \centering
    \caption{Image-text retrieval performance under different noise ratios on the RSICD dataset.}
\begin{tabular*}{\textwidth}{@{\extracolsep{\fill}}cccccccccc@{}}
\toprule
\multirow{2}{*}{Noise} & \multirow{2}{*}{Method} &\multirow{2}{*}{Ref.} & \multicolumn{3}{c}{Image-to-Text Retrieval} & \multicolumn{3}{c}{Text-to-Image Retrieval} & \multirow{2}{*}{mR}  \\
\cmidrule(lr){4-6} \cmidrule(lr){7-9}
 & & & R@1 & R@5 & R@10 & R@1 & R@5 & R@10
 \\
\midrule
\multirow{11}{*}{20\%} 
 & AMFMN&TGRS'22 & 4.68±0.33 & 15.08±0.48 & 24.99±1.12 & 4.08±0.10 & 16.31±0.64 & 27.65±0.83 & 15.47±0.34 \\
  & HVSA&TGRS'23 & 4.76±0.40 & 14.93±0.69 & 25.21±1.33 & 2.85±0.29 & 13.15±0.67 & 24.85±1.00 & 14.29±0.49 \\
  & KAMCL&TGRS'23 & 7.59±0.79 & 20.48±0.57 & 31.31±0.41 & 5.75±0.23 & 21.38±0.46 & 34.70±0.66 & 20.20±0.28 \\
  & SWAN&ICMR'23 & 4.50±0.62 & 13.76±0.82 & 22.71±1.04 & 3.45±0.21 & 15.43±0.85 & 28.54±0.69 & 14.73±0.25 \\
  & PIR&ACMMM'23 & 7.36±0.38 & 20.29±0.64 & 31.64±1.46 & 5.41±0.32 & 20.68±0.32 & 34.84±0.85 & 20.04±0.50 \\
  & S-CLIP&NeurIPS'23 & 3.82±0.27 & 14.84±1.39 & 26.26±1.16 & 3.90±0.33 & 15.13±1.22 & 27.19±1.20 & 15.19±0.67 \\
  & SIRS&TGRS'24 & 4.80±0.44 & 14.38±1.06 & 24.17±1.04 & 3.48±0.14 & 16.11±0.51 & 29.41±0.64 & 15.39±0.52 \\
  & MSA&TGRS'24 & 6.99±0.65 & 20.28±0.81 & 31.77±0.90 & 5.65±0.33 & 21.93±0.44 & 37.12±0.49 & 20.62±0.18 \\
  & SEMICLIP&ICLR'25 & 4.85±0.49 & 17.59±0.45 & 29.24±1.03 & 4.28±0.83 & 17.13±0.80 & 28.54±0.86 & 16.94±0.58 \\
  & CUP&TNNLS'25 & \underline{9.51±1.09} & \underline{29.29±1.05} & \underline{44.01±1.45} & \underline{7.45±0.44} & \underline{26.82±0.83} & \underline{42.29±0.94} & \underline{26.56±0.90} \\
  & RRSITR&Ours & \textbf{16.56±1.00} & \textbf{34.80±1.53} & \textbf{48.01±1.41} & \textbf{12.50±0.62} & \textbf{34.22±1.61} & \textbf{50.12±1.98} & \textbf{32.70±1.30} \\
\midrule
\multirow{11}{*}{40\%} 
 & AMFMN&TGRS'22 & 3.68±0.40 & 13.03±0.71 & 21.67±0.95 & 3.09±0.15 & 12.95±0.42 & 23.62±1.00 & 13.01±0.10 \\
  & HVSA&TGRS'23 & 3.47±0.30 & 13.32±0.51 & 22.16±0.96 & 1.82±0.30 & 9.84±0.32 & 18.34±0.35 & 11.49±0.22 \\
  & KAMCL&TGRS'23 & 3.49±0.74 & 10.50±1.55 & 17.80±1.59 & 2.41±0.66 & 11.11±2.00 & 21.06±2.91 & 11.06±1.32 \\
  & SWAN&ICMR'23 & 3.31±0.40 & 10.67±0.45 & 19.87±0.64 & 2.97±0.27 & 13.23±0.79 & 25.49±1.16 & 12.59±0.27 \\
  & PIR&ACMMM'23 & 4.61±0.10 & 14.99±0.79 & 24.67±0.64 & 3.80±0.38 & 16.48±0.29 & 29.53±0.44 & 15.68±0.25 \\
  & S-CLIP&NeurIPS'23 & 3.09±0.60 & 12.31±1.60 & 21.17±1.95 & 2.85±0.51 & 11.47±1.50 & 20.70±2.70 & 11.93±1.25 \\
  & SIRS&TGRS'24 & 3.28±0.34 & 11.36±0.37 & 20.38±1.00 & 2.90±0.24 & 13.14±0.68 & 24.07±0.36 & 12.52±0.41 \\
  & MSA&TGRS'24 & 2.19±0.46 & 7.63±1.59 & 13.83±2.14 & 3.27±0.33 & 14.04±0.56 & 26.00±1.16 & 11.16±0.96 \\
  & SEMICLIP&ICLR'25 & 3.66±0.32 & 13.70±0.94 & 22.58±1.02 & 2.91±0.62 & 11.91±1.02 & 20.62±1.50 & 12.56±0.77 \\
  & CUP&TNNLS'25 & \underline{9.11±0.83} & \underline{28.16±1.24} & \underline{42.16±1.57} & \underline{7.33±0.32} & \underline{24.83±0.73} & \underline{39.63±0.97} & \underline{25.20±0.59} \\
  & RRSITR&Ours & \textbf{15.77±0.70} & \textbf{34.15±0.42} & \textbf{47.36±1.53} & \textbf{11.81±0.88} & \textbf{35.09±0.65} & \textbf{50.90±1.38} & \textbf{32.51±0.63} \\
\midrule
\multirow{11}{*}{60\%} 
 & AMFMN&TGRS'22 & 3.44±0.55 & 10.87±0.62 & 18.87±1.41 & 2.49±0.27 & 10.75±0.67 & 19.29±1.41 & 10.95±0.56 \\
  & HVSA&TGRS'23 & 3.50±0.19 & 12.42±0.45 & 20.88±0.83 & 1.72±0.12 & 8.37±0.22 & 15.49±0.45 & 10.40±0.21 \\
  & KAMCL&TGRS'23 & 0.79±0.62 & 2.56±1.95 & 5.23±3.83 & 0.87±0.23 & 4.38±1.36 & 8.45±2.84 & 3.71±1.75 \\
  & SWAN&ICMR'23 & 2.80±0.21 & 8.84±1.04 & 16.72±1.93 & 2.82±0.28 & 12.88±0.65 & 23.95±0.98 & 11.34±0.41 \\
  & PIR&ACMMM'23 & 3.35±0.62 & 10.65±1.23 & 18.63±0.57 & 2.79±0.28 & 12.60±0.80 & 23.06±0.97 & 11.85±0.65 \\
  & S-CLIP&NeurIPS'23 & 0.46±0.28 & 1.85±0.68 & 3.40±0.69 & 0.16±0.10 & 1.10±0.32 & 2.32±0.79 & 1.55±0.42 \\
  & SIRS&TGRS'24 & 2.32±1.16 & 7.10±3.62 & 13.63±6.55 & 1.65±0.83 & 8.03±3.83 & 15.31±7.12 & 8.01±3.82 \\
  & MSA&TGRS'24 & 0.14±0.07 & 0.40±0.08 & 0.69±0.12 & 1.07±0.23 & 4.62±0.57 & 8.67±0.96 & 2.60±0.28 \\
  & SEMICLIP&ICLR'25 & 2.34±0.22 & 9.28±0.38 & 16.14±0.54 & 1.70±0.21 & 8.69±0.71 & 15.09±0.78 & 8.87±0.32 \\
  & CUP&TNNLS'25 & \underline{7.21±0.81} & \underline{24.58±0.80} & \underline{38.61±1.03} & \underline{6.21±0.57} & \underline{22.51±1.63} & \underline{36.34±1.71} & \underline{22.58±0.88} \\
  & RRSITR&Ours & \textbf{15.75±0.85} & \textbf{33.91±0.43} & \textbf{47.28±0.95} & \textbf{11.92±0.71} & \textbf{33.81±0.91} & \textbf{49.23±1.36} & \textbf{31.98±0.57} \\
\midrule
\multirow{11}{*}{80\%} 
 & AMFMN&TGRS'22 & 2.51±0.47 & 8.97±0.79 & 16.27±0.54 & 1.88±0.12 & 8.67±0.32 & 16.08±0.37 & 9.06±0.26 \\
  & HVSA&TGRS'23 & 1.98±0.22 & 8.91±0.56 & 15.33±1.26 & 0.96±0.10 & 5.02±0.44 & 9.43±0.91 & 6.94±0.48 \\
  & KAMCL&TGRS'23 & 0.11±0.04 & 0.25±0.04 & 0.51±0.09 & 0.09±0.00 & 0.45±0.06 & 1.00±0.14 & 0.40±0.05 \\
  & SWAN&ICMR'23 & 1.74±0.42 & 6.00±0.77 & 11.60±1.29 & 1.67±0.22 & 8.11±0.38 & 15.54±0.55 & 7.44±0.56 \\
  & PIR&ACMMM'23 & 2.49±0.50 & 9.57±0.78 & 18.41±1.39 & 2.10±0.12 & 10.80±0.80 & 20.46±0.97 & 10.64±0.59 \\
  & S-CLIP&NeurIPS'23 & 0.18±0.12 & 0.77±0.15 & 1.37±0.34 & 0.13±0.04 & 0.53±0.10 & 1.02±0.14 & 0.67±0.14 \\
  & SIRS&TGRS'24 & 1.86±0.34 & 6.13±0.57 & 11.18±0.89 & 1.11±0.05 & 5.15±0.28 & 9.57±0.47 & 5.83±0.38 \\
  & MSA&TGRS'24 & 0.09±0.00 & 0.46±0.13 & 0.73±0.15 & 0.31±0.10 & 1.84±0.10 & 3.31±0.34 & 1.12±0.10 \\
  & SEMICLIP&ICLR'25 & 1.21±0.17 & 4.37±0.84 & 7.67±0.96 & 0.82±0.39 & 4.12±1.01 & 7.12±1.68 & 4.22±0.78 \\
  & CUP&TNNLS'25 & \underline{5.87±0.56} & \underline{20.93±0.36} & \underline{34.33±0.82} & \underline{5.51±0.37} & \underline{19.70±0.93} & \underline{32.59±1.14} & \underline{19.82±0.45} \\
  & RRSITR&Ours & \textbf{13.47±0.90} & \textbf{28.80±1.47} & \textbf{41.41±1.81} & \textbf{10.02±1.04} & \textbf{30.55±1.70} & \textbf{45.13±1.93} & \textbf{28.23±1.26} \\
\bottomrule
\end{tabular*}
 \label{tab_rsicd} \vspace{-12pt}
\end{table*}
\begin{table*}[!htb]
    \scriptsize
    \setlength{\extrarowheight}{-1pt} 
    \centering
    \caption{Image-text retrieval performance under different noise ratios on the NWPU dataset.}
\begin{tabular*}{\textwidth}{@{\extracolsep{\fill}}cccccccccc@{}}
\toprule
\multirow{2}{*}{Noise} & \multirow{2}{*}{Method} &\multirow{2}{*}{Ref.} & \multicolumn{3}{c}{Image-to-Text Retrieval} & \multicolumn{3}{c}{Text-to-Image Retrieval} & \multirow{2}{*}{mR}  \\
\cmidrule(lr){4-6} \cmidrule(lr){7-9}
 & & & R@1 & R@5 & R@10 & R@1 & R@5 & R@10
 \\
\midrule
\multirow{11}{*}{20\%} 
 & AMFMN&TGRS'22 & 5.47±0.24 & 21.99±0.79 & 36.89±0.73 & 4.47±0.24 & 18.66±0.23 & 31.97±0.32 & 19.91±0.38 \\
  & HVSA&TGRS'23 & 2.80±0.34 & 10.97±0.34 & 19.92±0.50 & 1.68±0.07 & 7.96±0.20 & 15.50±0.17 & 9.81±0.18 \\
  & KAMCL&TGRS'23 & \underline{14.01±0.47} & \underline{43.13±0.61} & \underline{60.81±0.67} & \underline{9.12±0.15} & \underline{30.40±0.29} & \underline{46.23±0.36} & \underline{33.95±0.34} \\
  & SWAN&ICMR'23 & 2.42±0.20 & 11.43±0.59 & 20.98±0.41 & 2.72±0.12 & 11.85±0.23 & 21.69±0.49 & 11.85±0.20 \\
  & PIR&ACMMM'23 & 10.61±0.30 & 37.13±0.88 & 56.07±0.47 & 7.55±0.20 & 27.57±0.24 & 43.30±0.32 & 30.37±0.30 \\
  & S-CLIP&NeurIPS'23 & 2.73±0.13 & 11.85±0.47 & 21.19±0.58 & 2.86±0.14 & 11.77±0.48 & 21.19±0.53 & 11.93±0.28 \\
  & SIRS&TGRS'24 & 2.76±0.28 & 11.84±0.87 & 21.54±0.37 & 2.22±0.09 & 10.59±0.22 & 19.65±0.19 & 11.43±0.23 \\
& MSA&TGRS'24 & 6.46±0.35 & 24.39±0.90 & 40.02±0.94 & 6.69±0.22 & 24.96±0.61 & 40.33±0.89 & 23.81±0.47 \\
  & SEMICLIP&ICLR'25 & 5.66±0.48 & 16.81±0.54 & 28.81±0.70 & 4.43±0.23 & 17.26±0.71 & 29.41±0.49 & 17.06±0.37 \\
  & CUP&TNNLS'25 & 9.26±1.21 & 29.36±2.47 & 44.46±3.66 & 5.49±0.38 & 19.97±2.02 & 32.69±2.76 & 23.54±2.05 \\
  & RRSITR&Ours & \textbf{22.84±0.38} & \textbf{57.21±0.71} & \textbf{74.32±0.34} & \textbf{13.62±0.27} & \textbf{38.89±0.39} & \textbf{54.62±0.46} & \textbf{43.58±0.36} \\
\midrule
\multirow{11}{*}{40\%} 
 & AMFMN&TGRS'22 & 2.83±0.16 & 12.65±0.62 & 23.14±0.85 & 2.62±0.05 & 11.76±0.31 & 21.59±0.31 & 12.43±0.33 \\
  & HVSA&TGRS'23 & 0.14±0.19 & 0.62±0.84 & 1.20±1.67 & 0.14±0.22 & 0.65±0.98 & 1.23±1.73 & 0.66±0.94 \\
  & KAMCL&TGRS'23 & 2.58±0.91 & 11.99±3.34 & 21.16±5.77 & 3.25±0.48 & 13.90±1.58 & 24.13±2.44 & 12.84±2.27 \\
  & SWAN&ICMR'23 & 1.90±0.20 & 8.95±0.53 & 16.89±0.78 & 2.03±0.06 & 9.56±0.09 & 17.98±0.24 & 9.55±0.23 \\
  & PIR&ACMMM'23 &6.19±0.70 & 25.50±0.74 & \underline{41.58±1.09} & \underline{5.03±0.05} & \underline{19.98±0.17} & \underline{33.91±0.18} & \underline{22.03±0.43} \\
  & S-CLIP&NeurIPS'23 & 1.92±0.22 & 9.04±0.74 & 16.47±1.17 & 2.13±0.09 & 9.63±0.51 & 17.75±1.17 & 9.49±0.61 \\
  & SIRS&TGRS'24 & 2.27±0.13 & 10.40±0.26 & 18.75±0.66 & 2.10±0.06 & 9.84±0.19 & 18.52±0.22 & 10.31±0.24 \\
  & MSA&TGRS'24 & 2.10±0.19 & 9.27±0.72 & 17.34±1.05 & 2.12±0.11 & 9.82±0.35 & 18.75±0.45 & 9.90±0.34 \\
  & SEMICLIP&ICLR'25 & 4.47±0.23 & 13.01±0.37 & 22.44±0.46 & 3.16±0.23 & 13.27±0.19 & 23.14±0.33 & 13.25±0.16 \\
  & CUP&TNNLS'25 &  \underline{8.28±0.80} & \underline{27.11±2.80} & 41.52±3.80 & 4.87±0.62 & 17.79±1.70 & 29.69±2.42 & 21.54±1.98 \\
  & RRSITR&Ours & \textbf{20.55±0.32} & \textbf{53.89±0.63} & \textbf{71.10±0.53} & \textbf{12.56±0.16} & \textbf{36.67±0.17} & \textbf{52.51±0.45} & \textbf{41.21±0.19} \\
\midrule
\multirow{11}{*}{60\%} 
 & AMFMN&TGRS'22 & 2.15±0.12 & 9.12±0.76 & 17.00±1.30 & 1.94±0.03 & 9.18±0.14 & 17.35±0.28 & 9.46±0.39 \\
  & HVSA&TGRS'23 & 0.04±0.03 & 0.18±0.03 & 0.32±0.05 & 0.03±0.00 & 0.18±0.02 & 0.36±0.06 & 0.19±0.03 \\
  & KAMCL&TGRS'23 & 0.67±0.55 & 3.07±2.44 & 6.24±4.97 & 1.25±0.06 & 6.25±0.38 & 11.98±0.89 & 4.91±1.31 \\
  & SWAN&ICMR'23 & 1.42±0.09 & 7.47±0.21 & 14.27±0.36 & 1.77±0.11 & 8.47±0.25 & 15.95±0.36 & 8.23±0.06 \\
  & PIR&ACMMM'23 & 3.49±0.44 & 15.90±0.48 & 27.51±0.86 & 3.16±0.07 & 13.26±0.34 & 23.85±0.58 & 14.53±0.40 \\
  & S-CLIP&NeurIPS'23 & 0.14±0.06 & 0.75±0.30 & 1.36±0.45 & 0.13±0.06 & 0.69±0.26 & 1.28±0.45 & 0.73±0.25 \\
  & SIRS&TGRS'24 & 1.90±0.13 & 8.84±0.27 & 16.41±0.28 & 1.92±0.11 & 8.80±0.29 & 16.64±0.21 & 9.09±0.12 \\
  & MSA&TGRS'24 & 0.08±0.06 & 0.54±0.24 & 1.09±0.40 & 1.66±0.20 & 7.75±0.59 & 14.72±0.89 & 4.31±0.25 \\
  &SEMICLIP&ICLR'25 & 2.96±0.26 & 9.12±0.37 & 15.64±0.63 & 2.04±0.12 & 8.96±0.42 & 16.08±0.77 & 9.13±0.35 \\
  & CUP&TNNLS'25 & \underline{5.94±0.76} & \underline{20.05±1.33} & \underline{32.16±2.34} & \underline{3.58±0.36} & \underline{13.91±1.48} & \underline{23.90±2.07} & \underline{16.59±1.34} \\
  & RRSITR&Ours & \textbf{18.22±0.36} & \textbf{49.11±0.92} & \textbf{66.58±0.80} & \textbf{11.19±0.19} & \textbf{34.13±0.37} & \textbf{49.76±0.35} & \textbf{38.17±0.44} \\
\midrule
\multirow{11}{*}{80\%} 
 & AMFMN&TGRS'22 & 1.45±0.18 & 7.42±0.31 & 14.47±0.70 & 1.69±0.10 & 8.06±0.15 & 15.43±0.24 & 8.09±0.19 \\
  & HVSA&TGRS'23 & 0.02±0.02 & 0.19±0.07 & 0.34±0.13 & 0.04±0.01 & 0.16±0.03 & 0.34±0.05 & 0.18±0.03 \\
  & KAMCL&TGRS'23 & 0.02±0.01 & 0.12±0.03 & 0.27±0.04 & 0.02±0.01 & 0.17±0.01 & 0.31±0.01 & 0.15±0.01 \\
  & SWAN&ICMR'23 & 0.79±0.17 & 4.50±0.30 & 9.16±0.54 & 1.00±0.13 & 5.32±0.37 & 10.47±0.36 & 5.21±0.21 \\
  & PIR&ACMMM'23 & 1.70±0.25 & 7.83±0.44 & 15.25±0.64 & 1.44±0.03 & 7.03±0.12 &13.91±0.32 & 7.86±0.19 \\
  & S-CLIP&NeurIPS'23 & 0.02±0.01 & 0.17±0.01 & 0.32±0.00 & 0.03±0.00 & 0.17±0.01 & 0.33±0.01 & 0.17±0.00 \\
  & SIRS&TGRS'24 & 1.35±0.13 & 6.77±0.24 & 12.83±0.25 & 1.39±0.01 & 6.50±0.11 & 12.33±0.28 & 6.86±0.13 \\
  & MSA&TGRS'24 & 0.02±0.02 & 0.20±0.04 & 0.36±0.04 & 0.52±0.10 & 2.24±0.44 & 4.09±0.79 & 1.24±0.21 \\
  & SEMICLIP&ICLR'25 & 0.94±0.12 & 3.28±0.46 & 5.96±0.76 & 0.78±0.23 & 3.59±0.49 & 6.36±0.70 & 3.49±0.43 \\
  & CUP&TNNLS'25 & \underline{3.87±0.29} & \underline{14.50±1.23} & \underline{24.26±1.76} & \underline{2.35±0.17} & \underline{9.95±0.59} & \underline{18.30±0.90} & \underline{12.21±0.72} \\
  & RRSITR&Ours & \textbf{14.94±0.46} & \textbf{41.24±0.54} & \textbf{57.25±0.72} & \textbf{8.75±0.10} & \textbf{28.03±0.21} & \textbf{42.59±0.34} & \textbf{32.13±0.30}  \\
\bottomrule
\end{tabular*}
    \label{tab_nwpu}\vspace{-12pt}
\end{table*}
\subsection{Datasets}
In our experiments, to validate the effectiveness of the proposed method, we employ three benchmark datasets, i.e., RSITMD~\cite{yuan2022exploring}, RSICD~\cite{lu2017exploring}, and NWPU~\cite{cheng2022nwpu}. Following the protocol in~\cite{zhang2023hypersphere}, we split each dataset into 80\% for training, 10\% for validation, and 10\% for testing, respectively. To simulate real-world misalignment scenarios, we introduce synthetic noise into the training set by randomly shuffling text descriptions across images, thereby generating controlled NC. We employ four noise rates (i.e., 20\%, 40\%, 60\%, and 80\%) to evaluate model robustness under varying degrees of correspondence degradation. Dataset details and experimental results of 0\% noise rate are shown in supplementary materials\footnote{\url{https://github.com/MSFLabX/RRSITR}}.

\subsection{Implementation Details}
The RRSITR method adopts the ViT-B-32 architecture provided by OpenCLIP~\cite{cherti2023reproducible} as the backbone network. The model is trained using the Adam optimizer~\cite{kingma2015adam} for 50 epochs with a batch size of 100. We adopt a linear warm-up and cosine learning rate scheduler. Moreover, the learning rate, weight decay, warmup steps, and the maximum gradient norm are set to $7e^{-6}$, 0.7, 200, and 50, respectively. For some parameters, we set $\gamma_{1}$,  $\gamma_{2}$, $\sigma$, $\lambda_1$, $\lambda_2$, and $\alpha$ to 5, 18, 0.6, 0.8, 0.9, and 0.9 on all datasets. All experiments are implemented based on PyTorch and trained on a single NVIDIA A800 GPU.

\subsection{Comparison Methods}
We compare the proposed RRSITR with some state-of-the-art methods, including SIRS~\cite{zhu2024sirs}, MSA~\cite{yang2024transcending}, AMFMN~\cite{yuan2022exploring}, HVSA~\cite{zhang2023hypersphere}, SWAN~\cite{pan2023reducing}, PIR~\cite{pan2023prior}, and KAMCL~\cite{ji2023knowledge}, S-CLIP~\cite{mo2023s}, SEMICLIP~\cite{gan2025semisupervised}, and CUP~\cite{wang2025cross}. More details are shown in the supplementary materials. For all experiments, we utilize Rank at k (R@k, k=1,5,10) and mean Rank (mR) as the evaluation metrics to evaluate the retrieval performance~\cite{pan2023reducing}. Specifically, R@k measures the percentage of ground truth instances within the top R@k samples, while mR represents the average value across all R@k metrics, providing an overall assessment of retrieval performance. In our experiments, we select the best checkpoint according to validation performance. For fairness, we run five times using the recommended settings to record the mean scores and the standard deviation. For all experimental results, the highest scores and the second-highest scores are highlighted in \textbf{bold} and \underline{underlines}, respectively. 

\subsection{Comparison with the State-of-the-Art Methods}
Experimental results on the three datasets are reported in Tab.\ref{tab_rsitmd}, Tab.\ref{tab_rsicd}, and Tab.\ref{tab_nwpu}, respectively. From these results, we could observe that: 1) The RRSITR method demonstrates comprehensively leading performance and exceptional noise robustness across all three datasets. 2) As the noise rate increases sharply from 20\% to 80\%, the retrieval performance of all baseline methods declines substantially on the datasets with synthetic NC. However, our RRSITR maintains relatively high performance, with its stability far surpassing other methods under the extreme noise scenario (i.e., 80\% ). 3) In specific retrieval tasks, RRSITR achieves the highest scores on key metrics such as R@1, which stems from its cognitive learning strategy. Our RRSITR could effectively avoid the misleading of noisy pairs and enhance the distinction between positive and negative sample pairs.

\subsection{Ablation Studies}

To investigate the role of each component in our RRSITR, ablation studies are conducted on the RSITMD dataset with an 80\% noise ratio. More ablation studies are provided in the supplementary materials. We compare the proposed RRSITR with eight variants. Specifically, \#1, \#2, \#3, and \#4 represent the removal of the local contrastive learning module, SPL module, RTL module, and all components, respectively. To further validate the SPL and RTL modules, we replace the SPL module in RRSITR with \#5 (SPL with hard to easy strategy), \#6 (SPL with random sample weighting)  and \#7 (SPL with ambiguous category removal). Meanwhile‌, \#8 is defined by replacing the RTL module with fixed margin triplet loss. As shown in Tab.\ref{Ab}, we could obtain the following conclusion: 1) The complete RRSITR framework achieves the best overall performance under the 80\% noise condition, demonstrating the efficacy of each component. 2) The synergistic interactions among modules significantly enhance model robustness, effectively addressing noisy correspondence challenges. 3) Both the proposed multi-modal SPL and RTL maintain remarkable performance at the 80\% noise level, verifying their effectiveness and robustness in noisy environments.

\begin{table}[!htb]
    \centering
    
    \caption{Ablation studies on RSITMD with 80\% noise ratio.}
    \label{Ab}
    \scriptsize
    \setlength{\tabcolsep}{2pt}
    \begin{tabular}{@{}cccccccc@{}}
        \hline
        \scriptsize
        \multirow{2}{*}{Method} & \multicolumn{3}{c}{Image-to-Text Retrieval} & \multicolumn{3}{c}{Text-to-Image Retrieval} & \multirow{2}{*}{mR} \\
        \cmidrule(lr){2-4} \cmidrule(lr){5-7}
        & R@1 & R@5 & R@10 & R@1 & R@5 & R@10 & \\
        \hline
\#1 & \underline{15.44} & 31.64 & \underline{45.49} & 11.99 & 38.13 & 57.57 & 33.38 \\
\#2 & 12.08 & 27.79 & 39.42 & 10.29 & 35.47 & 53.73 & 29.80 \\
\#3 & 13.94 & 29.20 & 40.66 & 11.96 & 34.84 & 50.64 & 30.21 \\
\#4 & 13.76 & 30.22 & 43.36 & 11.88 & 37.50 & 56.72 & 32.24 \\
\#5 & 0.18 & 0.97 & 1.64 & 0.36 & 1.42 & 2.76 & 1.22 \\
\#6 & 10.58 & 25.31 & 36.11 & 9.36 & 31.22 & 46.53 & 26.52 \\
\#7 & 0.22 & 0.79 & 1.73 & 0.22 & 1.20 & 2.41 & 1.10 \\
\#8 & 15.09 & \underline{32.96} & \underline{45.49} & \underline{12.82} & \underline{39.45} & \underline{57.58} & \underline{33.90} \\
RRSITR& \textbf{16.90} & \textbf{33.98} & \textbf{46.82} & \textbf{13.72} & \textbf{42.70} & \textbf{61.47} & \textbf{35.93} \\
        \hline
    \end{tabular}\vspace{-15pt}
     
\end{table}

\begin{figure}[!ht]
    \centering
    \begin{minipage}{0.49\linewidth}
        \centering
        \includegraphics[width=\textwidth]{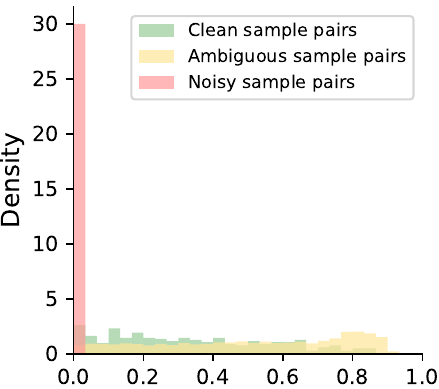}
    \end{minipage}
    \hfill
    \begin{minipage}{0.49\linewidth}
        \centering
        \includegraphics[width=\textwidth]{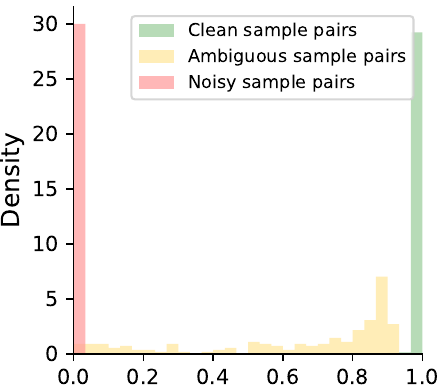}
    \end{minipage}
    \caption{\small Density versus weight of all sample pairs on RSITMD during training at the 1st epoch (left) and the 50th epoch (right).}
    \label{self_weight}\vspace{-15pt}
\end{figure}
\subsection{Self-paced Behavior Analysis}
To investigate the adaptive learning behavior of our proposed RRSITR model, we draw the density distribution of sample weights from different training epochs on the RSITMD dataset under a 40\% noise rate.  As shown in Fig.\ref{self_weight}, during the initial training phase, the RRSITR model first assigns zero weights to hard samples, treating them as noisy pairs. Thus, in the early stages of training, we can only better identify noisy pairs. As training progresses, we can effectively divide the training data into three subsets, i.e., clean, ambiguous, and noisy data. Our RRSITR gradually learns all clean and ambiguous samples from easy to hard, thereby enhancing its ability to distinguish noisy samples and improving the model's robustness to NC.

\begin{figure}[t]
\begin{minipage}{0.49\linewidth}
\centering
\includegraphics[width=\textwidth]{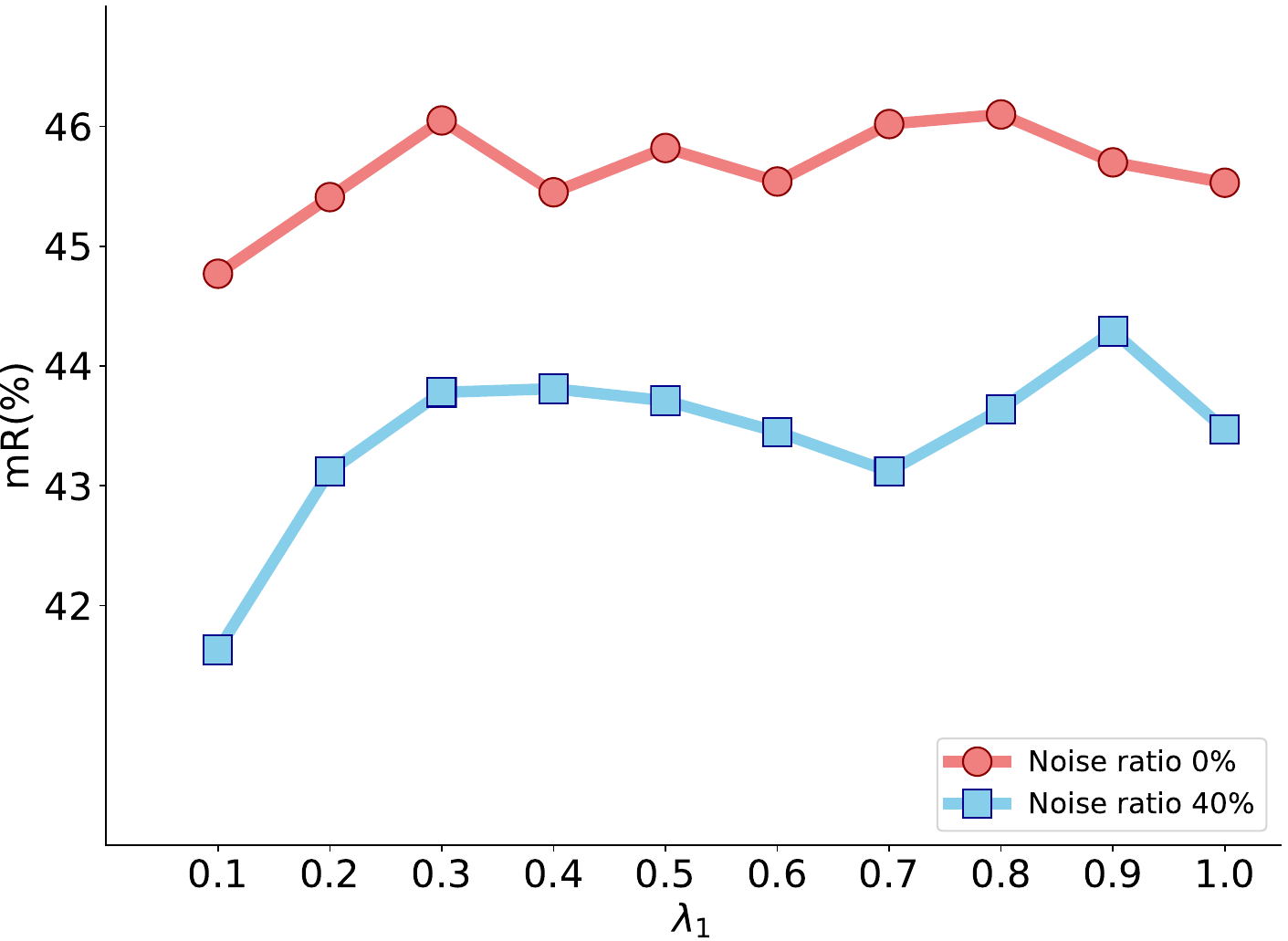}
\label{la}
\end{minipage}
\hfill
\begin{minipage}{0.49\linewidth}
\centering
\includegraphics[width=\textwidth]{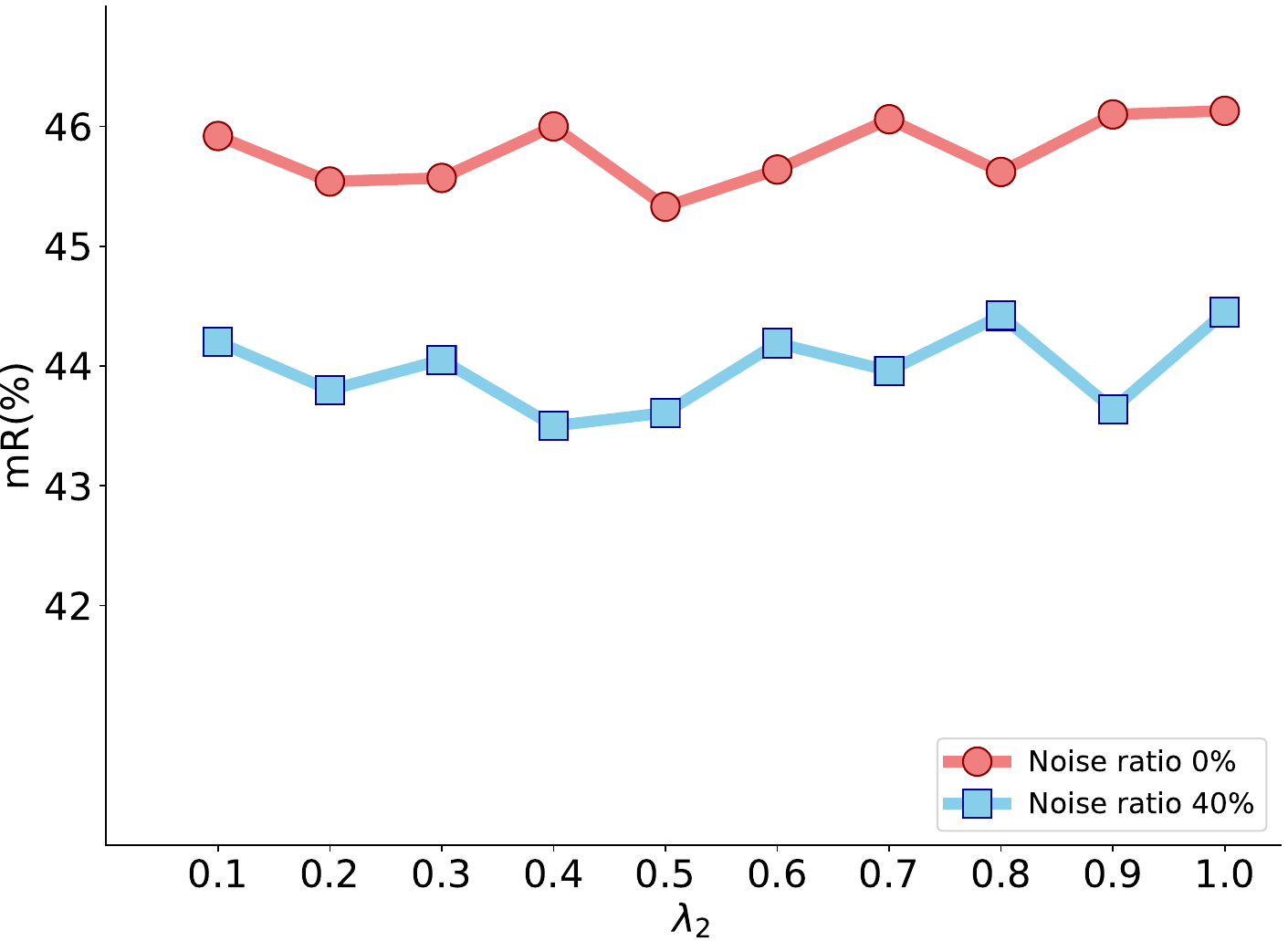}
\label{lb}
\end{minipage}
\vspace{-1.5em} 
\caption{\small Parameters analysis (i.e., $\lambda_1$ and $\lambda_2$) on RSITMD.} 
\label{pa}\vspace{-15pt}
\end{figure}

\subsection{Parameter Analysis}
This study investigates the effects of $\lambda_1$ and $\lambda_2$ on retrieval performance. More parameter experiments in the supplementary materials. As shown in Fig.\ref{pa}, too large or too small parameters will lead to poor performance. Specifically, we could observe that the model maintains satisfactory performance in both noisy and noise-free environments when $\lambda_1$ lies in the range of 0.7 to 0.9 and $\lambda_2$ lies in the range of 0.7 to 1.0. This demonstrates that such a parameter combination effectively balances the contributions of the self-paced loss and the robust triplet loss, thereby ensuring stable and superior performance under varying data quality conditions.

\section{Conclusion}
\label{s5}
This paper studies a challenging problem of Noisy Correspondence (NC) in RSITR, where mismatched image-text pairs are inevitably introduced into the training data, thereby leading to significant performance degradation. To address this issue, we propose a novel Robust Remote Sensing Image–Text Retrieval (RRSITR) paradigm. Specifically, our RRSITR first categorizes training samples into clean, ambiguous, and noisy pairs through fine-grained alignment. Then, we design a new self-paced learning function to progressively learn reliable image-text correspondences from easy to hard pairs by dynamically assessing the learning difficulty/reliability. Moreover, a robust triplet loss with dynamic soft margins is presented to reduce the negative influence of noisy pairs. Extensive experiments demonstrate that our proposed RRSITR could deal with NC and achieve superior performance compared to state-of-the-art methods under various noise rates.

\noindent\textbf{Acknowledgments} This work was supported in part by the National Natural Science Fund of China (Grant No. 62401204), the Hunan Provincial Natural Science Foundation of China under (Grant No. 2026JJ60221), the Sichuan Science and Technology Planning Project (Grant No. 2026NSFSC1480), the MoE Key Laboratory of Brain-inspired Intelligent Perception and Cognition, the University of Science and Technology of China, and the Open Project Program of the State Key Laboratory of CAD$\&$CG Zhejiang University (Grant No. A2512).


{
    \small
    \bibliographystyle{ieeenat_fullname}
    \bibliography{arxiv}
}


\end{document}